\documentclass[letterpaper]{article} 
\usepackage{aaai2026}  
\usepackage{times}  
\usepackage{helvet}  
\usepackage{courier}  
\usepackage[hyphens]{url}  
\usepackage{graphicx} 
\usepackage{natbib}  
\usepackage{caption} 
\usepackage{algorithm}
\usepackage{algorithmic}

\usepackage{newfloat}
\usepackage{listings}

\usepackage{multirow}
\usepackage{booktabs}

\usepackage{amsmath,amssymb}
\usepackage{xcolor}

\DeclareCaptionStyle{ruled}{labelfont=normalfont,labelsep=colon,strut=off} 
\floatstyle{ruled}
\newfloat{listing}{tb}{lst}{}
\floatname{listing}{Listing}
\title{Intention-Aware Diffusion Model for Pedestrian Trajectory Prediction}
\author{
    Yu Liu\textsuperscript{\rm 1,2}, 
    Zhijie Liu\textsuperscript{\rm 1}, 
    Xiao Ren\textsuperscript{\rm 1}, 
    You-Fu Li\textsuperscript{\rm 2},
    He Kong\textsuperscript{\rm 1},
}
\affiliations{
    \textsuperscript{\rm 1}Southern University of Science and Technology\\
    \textsuperscript{\rm 2}City University of Hong Kong\\
    yuliu254-c@my.cityu.edu.hk,\{12332642,12431359\}@mail.sustech.edu.cn,\\
    meyfli@cityu.edu.hk, kongh@sustech.edu.cn

}

\usepackage{bibentry}

\makeatletter
\gdef\copyright@on{} 
\makeatother

\begin{document}

\maketitle

\begin{abstract}
Predicting pedestrian motion trajectories is critical for the path planning and motion control of autonomous vehicles. Recent diffusion-based models have shown promising results in capturing the inherent stochasticity of pedestrian behavior for trajectory prediction. However, the absence of explicit semantic modelling of pedestrian intent in many diffusion-based methods may result in misinterpreted behaviors and reduced prediction accuracy. To address the above challenges, we propose a diffusion-based pedestrian trajectory prediction framework that incorporates both short-term and long-term motion intentions. Short-term intent is modelled using a residual polar representation, which decouples direction and magnitude to capture fine-grained local motion patterns. Long-term intent is estimated through a learnable, token-based endpoint predictor that generates multiple candidate goals with associated probabilities, enabling multimodal and context-aware intention modelling. Furthermore, we enhance the diffusion process by incorporating adaptive guidance and a residual noise predictor that dynamically refines denoising accuracy. The proposed framework is evaluated on the widely used ETH, UCY, and SDD benchmarks, demonstrating competitive results against state-of-the-art methods.

\end{abstract}


\section{Introduction}
Pedestrian motion prediction is a critical capability for extensive applications, including autonomous driving \cite{ref46,ref50,ref51}, robots navigation \cite{ref45} and planning \cite{ref47}. Given the observed trajectories of pedestrians, accurately forecasting their future paths is essential for ensuring safe and efficient operation. A major challenge lies in the inherently stochastic and non-deterministic nature of human motion, shaped by social interactions and environmental context. As pedestrian behavior unfolds over time, such influences manifest as both transient adjustments and global planning objectives. Addressing this challenge requires modelling both long-term destination goals and short-term motion intentions\cite{ref1, ref48}.

Within the above context, works such as \cite{ref30,ref37,ref38} adopt goal-based approaches, where trajectory endpoints are first predicted before inferring intermediate positions. This strategy aims to capture long-term dependencies and reveal the global motion tendencies. On the other hand, some methods incorporate midway points \cite{ref6,ref32,ref39}, which explicitly model the trajectory’s intermediate state to capture local dynamic patterns arising from fine-grained variations between consecutive timesteps. However, such anchor-based methods often rely on fixed spatial references to represent future motion, which may fail to capture the underlying semantics of pedestrian intent. These representations may misinterpret subtle variations in motion intention that are not truly indicative of a shift in the pedestrian's intended direction. For instance, a pedestrian slightly curving toward a building entrance may still maintain a steady forward-moving intent, whereas intermediate anchors might misleadingly indicate an unintended deviation or turning behavior.

\begin{figure}[t]
\centering
\includegraphics[scale= 0.70]{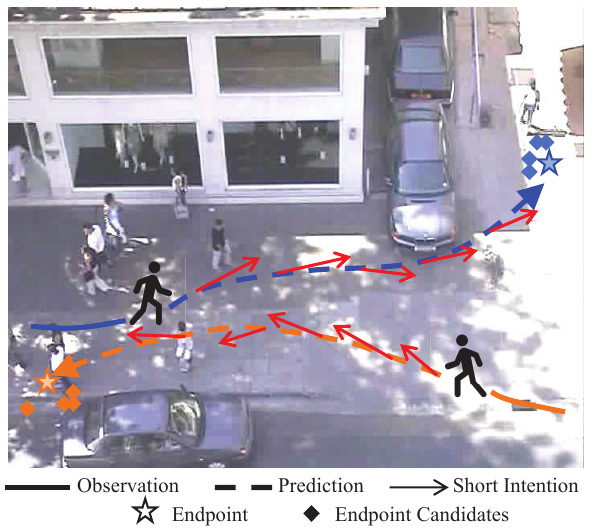}
\caption{An illustration of pedestrian intentions. The polar-based short-term intentions and long-term multimodal endpoints provide motion cues that guide the generation of future trajectories via a diffusion process.}
\label{intro}
\end{figure}

Another challenge lies in modelling pedestrian motion intentions. Some trajectory forecasting methods \cite{ref41, ref43, ref44} decompose intention into a set of discrete semantic categories, such as turning left, accelerating, or stopping. While this approach provides interpretability, it imposes rigid constraints on the inherently continuous and fine-grained nature of human movement. For example, a slight veer of 10 degrees and a sharp 90-degree turn may both be classified as turning left, despite their drastically different implications for trajectory evolution. Such categorical abstractions may oversimplify motion dynamics, making it difficult to capture the subtle variations and uncertainty that characterize real pedestrian behavior. These limitations highlight the need for a unified framework capable of capturing both the fine-grained nature of short-term motion and the uncertainty of long-term goals.

To tackle the above challenges, in this paper, we propose an Intention-Aware Diffusion model (IAD) for pedestrian trajectory prediction, which aims to capture the complicated long-term and short-term intention dependencies for the entire trajectory as shown in Figure \ref{intro}. In the short-term intention module, a residual polar representation is employed to model pedestrian motion intent. By separately encoding direction and magnitude, this formulation offers a compact, continuous, and expressive representation capable of capturing fine-grained motion patterns. The residual design promotes a structured modelling paradigm, where global motion cues provide a coarse prior that is refined through local adjustments, reflecting the hierarchical nature of human navigation. For long-term intention estimation, we propose a learnable token-based estimator that predicts multiple candidate endpoints with associated probabilities, capturing uncertainty and intent diversity. The token encodes trajectory-level context, enabling goal hypotheses that are both multimodal and context-aware, while the probabilistic formulation supports more informed and interpretable predictions. In the diffusion model, a residual noise prediction module is introduced to estimate the discrepancy between predicted and true noise, enabling dynamic refinement of the denoising process for improved trajectory generation. Further, an auxiliary guidance strategy enables smooth integration of conditional signals and balances diversity and fidelity.

\section{Related Works}

\subsection{Pedestrian Trajectory Prediction}
Early methods formulate predicting pedestrians' trajectories based on traditional ways, including social force \cite{ref25}, Kalman Filters \cite{ref26}, and Markov models \cite{ref27}. However, they may struggle to represent the complexity and variability of pedestrian motion in dynamic and crowded situations. Modern learning-based approaches have achieved significant advances. These methods model trajectory forecasting as a sequential estimation task using deep sequential processing models. For instance, Long Short-Term Memory (LSTM) and Recurrent Neural Networks (RNN) \cite{ref28, ref29} have been employed to capture pedestrian motion dynamics. Inspired by Natural Language Processing (NLP), Transformer models \cite{ref4, ref1} have been introduced to better capture long-range dependencies and global context, leading to improved performance in trajectory prediction tasks. Given the importance of social interactions, many approaches use graph-based \cite{ref11, ref6, ref2} structures to explicitly model complex inter-pedestrian relations.

\subsection{Diffusion Models for Trajectory Prediction}
Diffusion model \cite{ref14} is first proposed to solve non-equilibrium thermodynamics problems and have shown prior generation capabilities on various tasks, including image synthesis \cite{ref15, ref16}, video generation \cite{ref17, ref18}, and natural language processing \cite{ref19}. In the context of motion prediction, recent works have integrated diffusion models into prediction frameworks. MID \cite{ref7} explicitly simulates the process of human motion variation from indeterminate to determinate via the reverse process of diffusion. LED \cite{ref21} introduces a trainable leapfrog initialiser to skip denoising steps for prediction efficiency. TRACE \cite{ref22} constrains trajectories using target waypoints, speed, and specified social groups, while incorporating the surrounding context. DICE \cite{ref23} introduces an efficient sampling mechanism coupled with a scoring module to select the most plausible trajectories. C2F-TP \cite{ref24} proposes a coarse-to-fine prediction framework in which a conditional denoising model is used to refine the uncertainty of samples. However, these diffusion-based approaches do not consider motion intention in frameworks.  
\begin{figure*}[t]
\centering
\includegraphics[scale= 0.52]{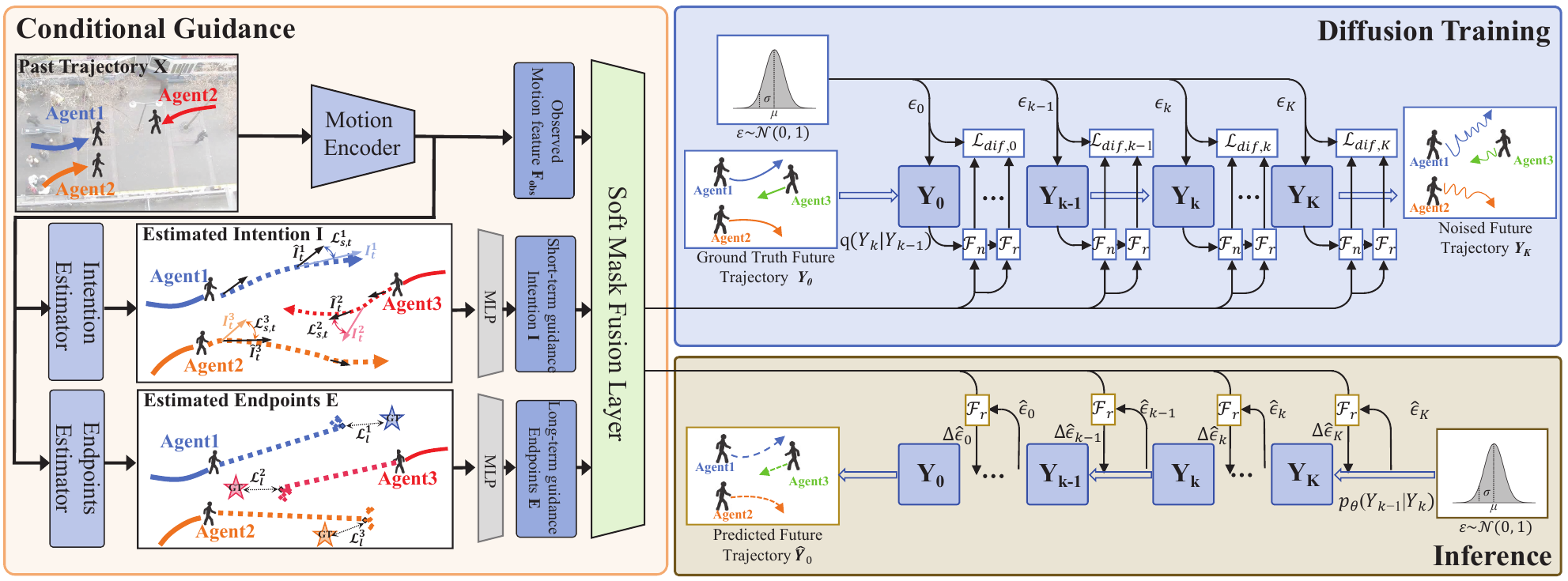}
\caption{The overall architecture of the proposed framework. The conditional guidance module estimates both short-term and long-term motion intentions, which are integrated through a soft-mask fusion layer. In the diffusion module, the noise estimator $\mathcal{F}_n$ works in tandem with the refinement network $\mathcal{F}_r$ to iteratively denoise and generate future trajectories.}
\label{structure1}
\end{figure*}

\subsection{Prior Conditioned Approach}
Due to the stochastic nature of human motion, pedestrians' trajectories contain randomness. To enable controllable and interpretable trajectory predictions, some approaches incorporate anchor points as prior information to guide the generation of multimodal trajectories. TNT \cite{ref30} formulates trajectory prediction as a two-stage process by first predicting discrete future endpoints and then generating target-conditioned trajectories. DenseTNT \cite{ref31} builds on this by replacing discrete endpoint classification with continuous density regression, improving spatial coverage and prediction accuracy.\cite{ref33} generates diverse proposals fused with goal-oriented anchors to enable multimodal prediction. Graph-TERN \cite{ref6} predicts intermediate control points by segmenting the future path, and refines trajectories using a spatiotemporal multi-relational graph for better accuracy. PPT \cite{ref1} progressively trains the model through next-step prediction, destination prediction, and full trajectory prediction, capturing both short- and long-term motion patterns. SingularTrajectory \cite{ref32} proposes an adaptive method which corrects misplaced anchors based on a traversability map. However, lacking semantic intent modelling may hinder prediction quality.  


\section{Methods}
\subsection{Preliminaries}
The diffusion model comprises a forward process and a reverse process. In the forward process, Gaussian noise is incrementally added to a sample drawn from the data distribution $\{Y_k\}^{K}_{k=0} \in \mathbb{R}^{t_{pred}\times2}  $, which corresponds to the future trajectories. This process starts from ground truth path $Y_0$ and is repeated for $K$ steps following a predefined noise schedule, gradually transforming the sample into a standard Gaussian distribution, which is mathematically defined : 
\begin{equation}
    q(Y_{k}{\mid}Y_{k-1} ) = \mathcal{N}(Y_{k}; \sqrt{1-\beta_{k}}Y_{k-1}, \beta_{k}I),
\end{equation}
where $\beta_k \in (0,1)$ denotes the rescaled variance schedule that controls the magnitude of noise added at each step. To reduce the computational cost during training, this process can be simplified using properties of Gaussian transitions:
\begin{equation}
    \begin{array}{cc}
        q(Y_{k}{\mid}Y_{0} ) = \mathcal{N}(Y_{k}; \sqrt{\bar{\alpha}_{k}}Y_{0}, (1-{\bar{\alpha}_{k}})I), 
    \end{array}
\end{equation}
\begin{equation}
    \begin{array}{cc}
        Y_{k} = \sqrt{\bar{\alpha}_{k}}Y_{0} + \sqrt{(1-{\bar{\alpha}_{k}})}\epsilon,\  \epsilon \sim \mathcal{N}(0, I),
    \end{array}
\end{equation}
where $\alpha_k = 1 - \beta_k$, $\bar{\alpha}_k = \prod_{i=1}^{k} \alpha_i$, and $\epsilon$ represents a noise vector sampled from a standard Gaussian distribution. As the diffusion step $k$ increases, $Y_k$ progressively approximates the standard normal distribution $\mathcal{N}(0, I)$.

The reverse process aims to remove noise from $Y_k$, gradually recovering the original distribution $Y_0$ by learning reverse $p_{\theta}(Y_{k-1}|Y_{k})$, which can be formulated as:
\begin{equation}
    p_{\theta}(Y_{k-1}{\mid}Y_{k} ) = \mathcal{N}(Y_{k-1}; \mu_{\theta}(Y_k, k,f), \Sigma_{\theta}(Y_k, k,f)),
\end{equation}
where $\mu_{\theta}$ and $\Sigma_{\theta}$ are neural networks predicting the mean and variance, respectively, and $f$ is the conditional guidance.

\subsection{Problem Definition}
The trajectory prediction task entails estimating pedestrians’ future positions within a scene based on their observed past movements. The model takes as input the 2D spatial coordinates of pedestrians over the observed time steps, denoted by $X_{1:t_{obs}} = \{X^i_t \in \mathbb{R}^2 | 1 \leq t \leq t_{obs} \} $ , where $X^i_t = (x^i_t, y^i_t)$ represents the position of the $i$th pedestrian at time $t$. Similarly, the ground truth trajectory over the future time period is denoted as $Y_{1:t_{pred}} = \{Y^i_t \in \mathbb{R}^2 | 1 \leq t \leq t_{pred} \}$, and $Y_t^i = (x_t^i, y_t^i)$ represents the ground truth position of the $i$th pedestrian at time $t$ in the future. The objective of this work is to predict the future trajectory $\hat{Y}^i$ of the $i$th pedestrian and to estimate its future positions $\hat{Y}^i_t = (\hat{x}^i_t, \hat{y}^i_t)$ at each time $t$.

\subsection{Overview}
The proposed architecture, illustrated in Figure~\ref{structure1}, is a diffusion-based framework that models both long-term $E$ and short-term $I$ pedestrian motion intentions through dedicated modules with observed motion features $F_{obs}$ as conditional guidance. During the trajectory generation process, a softmask classifier-free guidance mechanism is employed to adaptively integrate conditional signals and a refinement module estimates the residual error in the predicted noise.

\subsection{Motion Encoder}
The motion encoder extracts motion features from observed pedestrian trajectories while capturing social interactions. These features are used both for predicting intentions and conditioning the diffusion model for trajectory generation.
\begin{equation}
F_{obs} = \text{MotionEncoder}(X_{1:t_{obs}}) \in \mathbb{R}^{t_{obs} \times d},
\end{equation}
where $X \in \mathbb{R}^{t_{obs} \times 2}$ denotes the observed trajectories. We adopt the encoder from \cite{ref7}, proven effective in capturing complex pedestrian motion patterns.

\subsection{Residual Polar Modeling for Intention Estimation}
\begin{figure}[t]
\centering
\includegraphics[scale= 0.45]{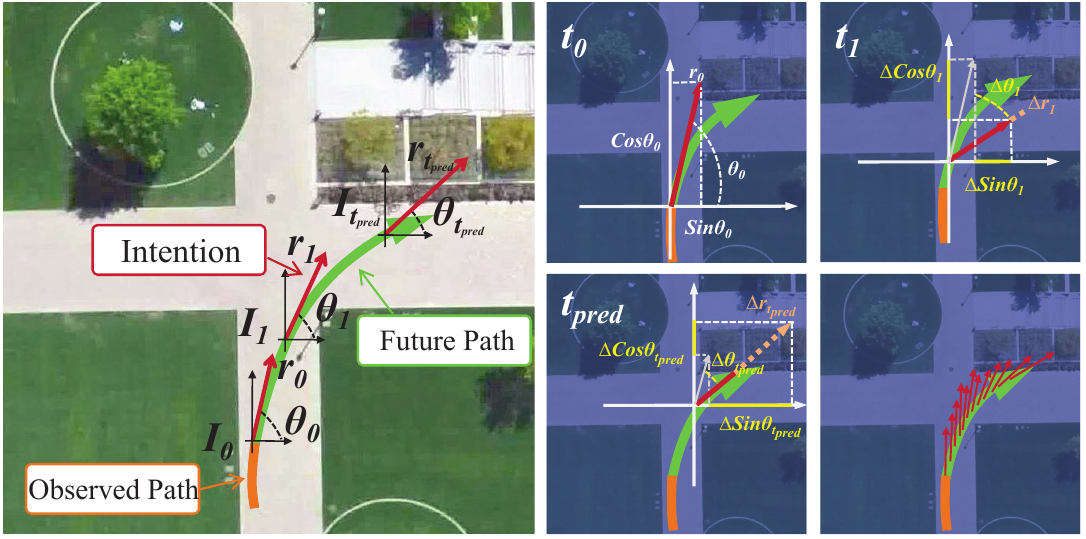}
\caption{Illustration of short-term intention prediction. The sequence of short-term intentions is constructed using a polar-coordinate representation, as shown on the left. On the right, residual updates are recursively accumulated over time to refine these intention predictions.}
\label{structur2}
\end{figure}

We propose a residual polar coordinate-based representation to model short-term pedestrian motion intention, inferred from observed motion features as shown in Figure \ref{structur2}. Instead of discretizing pedestrian intent into predefined motion classes (e.g., turning left or accelerating), intention at each future timestep $t$ is represented in a continuous polar form as $I^i = \{ \theta_t^i, r_t^i \}_{t=1}^{T_\mathrm{pred}} \in \mathbb{R}^{T_\mathrm{pred} \times 2}$, where $\theta_t^i \in (-\pi, \pi]$ denotes the directional angle, and $r_t^i \in \mathbb{R}^+$ denotes the magnitude of motion tendency. The origin point $\mathcal{O}_t^i$ corresponds to the pedestrian’s current Cartesian position $\{ x_i^t, y_i^t \}$. These components are computed as follows:
\begin{equation}
\theta^{i}_t = \arctan2(a^{y_i}_t, a^{x_i}_t), r^{i}_t = \sqrt{(a^{x_i}_t)^2 + (a^{y_i}_t)^2 + \varepsilon},
\end{equation}
where $\arctan2$ is the two-argument inverse tangent function, and $a^{x_i}_t$, $a^{y_i}_t$ denote the second-order temporal derivatives of the $i$th pedestrian's position in the lateral and longitudinal directions. A small constant $\varepsilon$ is added to ensure numerical stability and avoid division by zero.

To capture the dynamic changes in future motion intentions, we adopt a transformer network to predict a sequence $I^i$ of polar-coordinate representations. Instead of directly regressing the absolute values of the heading angle $\theta^{i}_t$ and magnitude $r^{i}_t$ at each time step, we reformulate the prediction task into estimating the residual changes with respect to the previous state. This is motivated by predicting residuals simplifies the learning objective by focusing on local variations, which are typically smoother and less noisy than absolute values. The process is defined as:
\begin{equation}
\left[\Delta \cos{\theta^{i}}, \Delta \sin{\theta^{i}}, \Delta r^{i} \right] = \text{IntentPredictor}(F_{obs}).
\end{equation}

We then convert the predicted residuals into scalar angular and magnitude increments, respectively: $\Delta \theta^{i}_t = \arctan2(\Delta \sin{\theta^{i}_t}, \Delta \cos{\theta^{i}_t})$ and $\Delta r^{i}_t = \mathrm{softplus}(\Delta r^{i}_t) = \log(1 + \exp(\Delta r^{i}_t))$. The residual polar updates for the future intention $I^i$ at each time step $t$ are then computed by recursively accumulating these residuals over time:
\begin{equation}
\theta^{i}_t = \theta^{i}_0 + \sum\nolimits_{\tau=1}^{t} \Delta \theta^{i}_\tau, \quad
r^{i}_t = r^{i}_0 + \sum\nolimits_{\tau=1}^{t} \Delta r^{i}_\tau,
\end{equation}
where $\theta^{i}_0$ and $r^{i}_0$ are initialised values based on the final frame of the observed trajectory $X^i_{t_{\mathrm{obs}}} =\{x^i_{t_{obs}}, y^i_{t_{obs}}\} $. 

This formulation incrementally refines intention predictions, ensuring smooth, consistent direction and magnitude estimates that better capture local motion dynamics.

\subsection{Endpoints Prediction}

\begin{figure}[t]
\centering
\includegraphics[scale= 0.70]{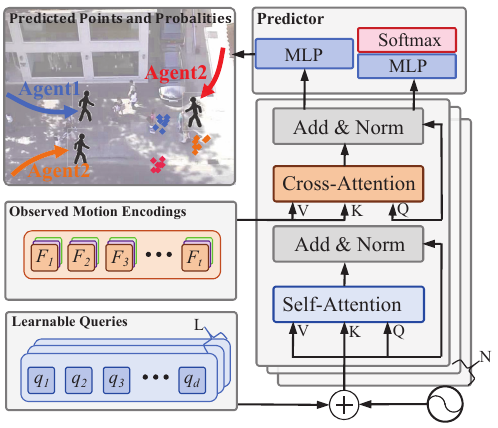}
\caption{Illustration of the Endpoint Estimator. $L$ learnable query tokens interact with observation features through cross-attention to generate multimodal endpoint predictions and associated scores. }
\label{structur3}
\end{figure}

The objective of destination prediction is to estimate the final endpoint of a pedestrian’s trajectory, thereby capturing long-term behavioral intent based on observed motion history. To address the inherent multimodality of human behavior, our model predicts multiple candidate destinations. During training, the candidate closest to the ground truth is selected as supervision, while during inference, the destination with the highest confidence score is chosen.

In this task, we use a learnable query-based method with a transformer block to generate diverse destinations as shown in Figure \ref{structur3}. Specifically, we initialize $L$ learnable endpoint query tokens $Q \in \mathbb{R}^{L \times d_{q}}$. These queries act as anchors representing plausible motion modes. To condition the queries on observed behavior, we refine them via the transformer’s cross-attention using motion features $F_{obs} \in \mathbb{R}^{T_{\text{obs}} \times d}$.

For endpoint predictor, an MLP is applied to the refined multimodal motion features obtained from the transformer to regress a set of candidate endpoints, denoted as $E^i = \{ e^{i}_1, e^{i}_2, \dots, e^{i}_L \} \in \mathbb{R}^{L \times 2}$. Additionally, a separate MLP followed by a softmax layer is used to predict a confidence score for each candidate, forming a discrete probability distribution over the endpoint hypotheses: $P^{i} = \{ p^{i}_1, p^{i}_2, \dots, p^{i}_L \} \in \mathbb{R}^{L \times 1}$, where $\sum_{l=1}^{L} p^{i}_l = 1$. The endpoint prediction module is given as:
\begin{equation}
\left[ E, P \right] = \text{EndpointPredictor}(F_{obs}, Q),
\end{equation}
where $\text{EndpointPredictor}$ is conditioned on observed motion features $F_{obs}$ and learnable goal queries $Q$.

\subsection{Condition-Guided Refinement }

To guide the generation process toward intended motion semantics, we first adopt a modified classifier-free guidance strategy with dynamic masking, followed by a residual refinement step to correct prediction errors.

Instead of statically injecting conditions, we propose a dynamic soft-mask mechanism to modulate the contribution of each guidance signal. Specifically, a multi-layer perception layer followed by a sigmoid activation is used to learn a soft weight $M_m$ for each guidance source $G_m$, where $m \in \{o, s, l\}$ denotes the modality source: observed trajectory feature $o$, short-term intention $s$, and long-term goal $l$.
\begin{equation}
M_{m} = \text{Sigmoid(}\text{MLP}_m(G_m)),
\end{equation}
\begin{equation}
G^{\prime}_{m} = (1 - M_m) \cdot G_m + M_m \cdot \psi_m,
\end{equation}
where $\psi_m$ denotes a learnable token for masked guidance.

These soft-masked features $G^{\prime}_m$ are concatenated with the noised trajectory state $Y_k$ and the corresponding diffusion step embedding $E_k$, then used to predict noise $\epsilon_{k} \in \mathbb{R}^{t_{obs} \times 2}$.
\begin{equation}
{\epsilon}_{k} = \text{NoiseEstimator}(\text{Concat}(G^{\prime}_o, G^{\prime}_s, G^{\prime}_l, Y_k, E_k)).
\end{equation}

Although the denoising network predicts the added noise ${\epsilon}_{k}$, directly learning the full noise can be suboptimal due to its inherent stochasticity. To address this, we introduce a residual refinement module that explicitly learns to correct the residual error $\Delta \epsilon_{k}$ not captured by the main diffusion model. The final refined noise is computed as:
\begin{equation}
{\epsilon}_{\text{refined},k} = {\epsilon}_{k} + \Delta \epsilon_{k},
\end{equation}
where $\Delta \epsilon_{k} = \text{RefineNet}(G^{\prime}_{o}, G^{\prime}_{s}, G^{\prime}_{l}, {\epsilon}_{k}, E_k) \in \mathbb{R}^{t_{obs} \times 2}$ denotes the predicted residual by the refinement network.

\subsection{Training Optimization}
To train the proposed method, we incorporate the short-term loss and the long-term loss in addition to the diffusion loss.

The short-term loss $\mathcal{L}_s$ supervises the intention estimator to more accurately approximate pedestrian motion tendencies within the observation context. Given the polar representation $(\theta, r)$ of intention, the angular component is trained using cosine distance to address its circular nature, while the magnitude is optimized via mean squared error.
\begin{equation}
    \mathcal{L}_\theta = \frac{1}{T 
    \times N} \sum_{i=1}^{N}\sum_{t=1}^{T} \left(1 - \cos(\hat{\theta}_{t}^{i} - \theta_{t}^{i}) \right),
\end{equation}
\begin{equation}
    \mathcal{L}_r = \frac{1}{T \times N} \sum_{i=1}^{N} \sum_{t=1}^{T} \, \text{MSE}\left(\hat{r}_{t}^{i} - r_{t}^{i} \right),
\end{equation}
\begin{equation}
    \mathcal{L}_{s} = \lambda_\theta \cdot \mathcal{L}_\theta + \lambda_r \cdot \mathcal{L}_r,
\end{equation}
where $\hat{\theta}^{i}_{t}$ and $\hat{r}^{i}_{t}$ are the predicted values, and ${\theta}^{i}_{t}$ and ${r}^{i}_{t}$ are the corresponding ground truths. The loss weights $\lambda_\theta$ and $\lambda_r$ are empirically set to 0.5 and 0.25.

The long-term loss guides the endpoint estimator to capture global motion patterns and accurately predict pedestrian destinations. Rather than optimizing over all candidate endpoints, which could dilute the learning signal due to the inherently multimodal nature of human motion, the endpoint loss $\mathcal{L}_{e}$ focuses on only minimizing the distance between the ground-truth endpoint and the closest predicted candidate.
\begin{equation}
    \mathcal{L}_{e} = \frac{1}{N} \sum_{i=1}^{N} \min_{j \in L} \, \text{MSE}(\hat{e}^i_j,e^i),
\end{equation}
where $\hat{e}^i$ and $e^i$ denote the predicted and ground-truth endpoints, respectively.

Additionally, a negative log-likelihood loss $\mathcal{L}_p$ boosts the confidence of the closest goal, encouraging high probability for accurate predictions. To suppress incorrect candidates and enforce a sharper distribution, a penalty term is added.
\begin{equation}
    \mathcal{L}_p = \frac{1}{N} \sum_{i=1}^{N}(-\log \hat{p}^i_{l^*} + \sum_{\substack{l \ne l^*}} \log \hat{p}^i_l ),
\end{equation}
\begin{equation}
    \mathcal{L}_{l} = \lambda_e \cdot \mathcal{L}_e + \lambda_p \cdot \mathcal{L}_p,
\end{equation}
where $\hat{p}^i$ is the predicted probability over endpoint candidates, and $l^*$ indicates the index of the predicted endpoint closest to the ground truth. The loss weights $\lambda_e$ and $\lambda_p$ are set to 1.0 and 0.5, respectively.

The diffusion loss effectively guides the noise estimator to generate reliable predictions $Y_k$ for use during inference, and is subsequently integrated with both the short intention and goal objectives to form the final weighted training loss.
\begin{equation}
\mathcal{L}_{dif} = \mathbb{E}_{k} \left\| \epsilon_{k} - \hat{\epsilon}_{refined,k} \right\|^2,
\end{equation}
\begin{equation}
    \mathcal{L} = \mathcal{L}_{int} + \mathcal{L}_{end} + \lambda_{dif} \cdot \mathcal{L}_{dif},
\end{equation}
where the diffusion weight $\mathcal{L}_{\text{dif}}$ is empirically set to 1.


\subsection{Inference}
The reverse diffusion process is typically modeled as a conditional Markov chain, which gradually denoises inputs to generate trajectories. However, its stochasticity and iterative nature incur high computational costs and may cause deviations from ground truth. To mitigate this, we adopt the deterministic DDIM sampling strategy, which removes randomness and reduces inference steps. Under this method, a sample $Y_{k-\gamma}$ is deterministically derived from $Y_k$ as:
\begin{equation}
\begin{aligned}
           Y_{k-\gamma} &= \sqrt{\alpha_{k-\gamma}} \left( \frac{Y_k - \sqrt{1 - \alpha_k} \, \epsilon_{refined}}{\sqrt{\alpha_k}} \right) \\
           &  + \sqrt{1 - \alpha_{k-\gamma}} \, \epsilon_{refined}.
\end{aligned}
\end{equation}

\begin{table*}[t]
\centering
\setlength{\tabcolsep}{7pt}  
\fontsize{10}{10}\selectfont
\begin{tabular}{lcc|ccccccccc}
\toprule
Method & Venue & Year & ETH & HOTEL & UNIV & ZARA1 & ZARA2 & AVG \\
\midrule
Social-LSTM       & CVPR & 2016 & 1.09 / 2.35 & 0.79 / 1.76 & 0.67 / 1.40 & 0.47 / 1.00 & 0.56 / 1.17 & 0.72 / 1.54 \\
Social-GAN        & CVPR & 2018 & 0.87 / 1.62 & 0.67 / 1.37 & 0.76 / 1.52 & 0.35 / 0.68 & 0.42 / 0.84 & 0.61 / 1.21 \\
Social-STGCNN     & CVPR & 2020 & 0.64 / 1.11 & 0.49 / 0.85 & 0.44 / 0.79 & 0.34 / 0.53 & 0.30 / 0.48 & 0.44 / 0.75 \\
Social-VAE        & ECCV & 2022 & 0.41 / 0.58 & 0.13 / 0.19 & \underline{0.21} / \underline{0.36} & \underline{0.17} / 0.29 & 0.13 / 0.22 & 0.21 / 0.33 \\
MID               & CVPR & 2022 & 0.39 / 0.66 & 0.13 / 0.22 & 0.22 / 0.45 & \underline{0.17} / 0.30 & 0.13 / 0.27 & 0.21 / 0.38 \\
Graph-TERN        & AAAI & 2023 & 0.42 / 0.58 & 0.14 / 0.23 & 0.26 / 0.45 & 0.21 / 0.37 & 0.17 / 0.29 & 0.24 / 0.88 \\
EigenTrajectory   & ICCV & 2023 & \underline{0.36} / 0.53 & \underline{0.12} / 0.19 & 0.24 / 0.43 & 0.19 / 0.33 & 0.14 / 0.24 & 0.21 / 0.34 \\
TUTR              & ICCV & 2023 & 0.40 / 0.61 & \textbf{0.11} / 0.18 & 0.23 / 0.42 & 0.18 / 0.34 & 0.13 / 0.25 & 0.21 / 0.36 \\
SMEMO             & TPAMI& 2024 & 0.39 / 0.59  & 0.14 / 0.20 & 0.23 / 0.41 & 0.19 / 0.32 & 0.15 / 0.26 & 0.22 / 0.35 \\
HighGraph         & CVPR & 2024 & 0.40 / 0.55 & 0.13 / \underline{0.17} & \textbf{0.20} / \textbf{0.33} & \underline{0.17} / \underline{0.27} & \textbf{0.11} / \underline{0.21} & \underline{0.20} / \textbf{0.30} \\
PPT               & ECCV & 2024 & \underline{0.36} / \textbf{0.51} & \textbf{0.11} / \textbf{0.15} & 0.22 / 0.40 & \underline{0.17} / 0.30 & \underline{0.12} / \underline{0.21} & \underline{0.20} / \underline{0.31} \\
\midrule
Ours              & --   & --   & \textbf{0.34} / \underline{0.52} & 0.15 / 0.24 & \textbf{0.20} / \underline{0.36} & \textbf{0.15} / \textbf{0.24} & \textbf{0.11} / \textbf{0.20} & \textbf{0.19} / \underline{0.31} \\
\bottomrule
\end{tabular}
\caption{Quantitative comparisons with state-of-the-art methods on the ETH/UCY dataset. Bold numbers indicate the best performance. Underlined numbers denote the second-best results.
}
\label{result1}
\end{table*}

\section{Experiments}
\subsection{Experimental Settings}

\subsubsection{Datasets:} In this study, we evaluate the proposed model on three widely used pedestrian trajectory datasets: ETH \cite{ref12}, UCY \cite{ref13}, and the Stanford Drone Dataset (SDD) \cite{ref49}. SDD is a large-scale dataset captured from a bird’s-eye view using drone cameras. The ETH dataset comprises two scenes, ETH and HOTEL, while the UCY dataset includes three scenes: ZARA1, ZARA2, and UNIV. All trajectories are provided in world coordinates, and evaluation results are reported in meters. All sequences consist of 8 observed and 12 predicted frames over 8 seconds.

\subsubsection{Metrics:} Following previous works, we adopt two standard evaluation metrics to assess the performance of the proposed method. Average Displacement Error (ADE) measures the mean Euclidean distance between the predicted and ground-truth trajectories over all prediction time steps, while Final Displacement Error (FDE) computes the Euclidean distance between the predicted final position and the ground-truth endpoint. In line with prior studies, we report the best result among 20 sampled trajectories.

\subsubsection{Implementation:} The network is implemented in PyTorch. The noise estimation module uses 4 Transformer layers with hidden size 512 and 4 attention heads. The short-term intention estimator employs 4 self-attention layers mapping features to 256 dimensions, while the long-term endpoint estimator adopts a similar structure with cross-attention to capture trajectory-level context. We set diffusion steps to $K=100$ and apply DDIM sampling with stride 20. The model is trained using Adam with learning rate 0.001 and batch size 256. All experiments are conducted on NVIDIA RTX 5090 GPUs and Intel Xeon 8481C CPUs.

\subsection{Quanitative Evaluation}
The methods compared in this work include: PPT \cite{ref1}, HighGraph \cite{ref2}, SMEMO \cite{ref3}, TUTR \cite{ref4}, EigenTrajectory \cite{ref5}, Graph-TERN \cite{ref6}, MID \cite{ref7}, Social-VAE \cite{ref8}, Social-STGCNN \cite{ref9}, Social-GAN \cite{ref10}, Social-LSTM \cite{ref28}, SOPHIE \cite{ref34}, PECNet \cite{ref35}, and PCCSNet \cite{ref36}.

The quantitative evaluation results on the UCY/ETH datasets are summarized in Table \ref{result1}, comparing our proposed approach against existing methods. Despite variations across individual subsets, our method consistently delivers competitive performance. Specifically, it achieves the lowest ADE on 4 out of 5 datasets and ranks either first or second in FDE on 4 of them. On average, the ADE is reduced from 0.20 to 0.19. In the ETH scenario, our method attains the best ADE, lowering it from 0.36 to 0.34. Similarly, for the Zara1 subset, the ADE is reduced from 0.17 to 0.15, and the FDE from 0.27 to 0.24. As shown in Table \ref{result2}, the proposed method achieves competitive results on the SDD dataset, with ADE reduced from 7.03 to 6.85 and the second-best FDE.

\begin{table}[t]
\centering
\setlength{\tabcolsep}{5pt}
\begin{tabular*}{\linewidth}{@{\extracolsep{\fill}}lccc c}
\toprule
Methods & Venue & Year & ADE & FDE \\
\midrule
Social-GAN   & CVPR & 2018 & 27.23 & 41.44 \\
Sophie       & CVPR & 2019 & 16.27 & 29.38  \\
PECNet       & ECCV & 2020 & 9.96 & 15.88 \\
PCCSNet      & ICCV & 2021 & 8.62 & 16.16 \\
Social-VAE   & ECCV & 2022 & 8.10 & 11.72 \\
MID          & CVPR & 2022 & 7.61 & 14.30 \\
LED          & CVPR & 2023 & 8.48 & 11.66 \\
TUTR         & ICCV & 2023 & 7.76 & 12.69 \\
PPT          & ECCV & 2024 & \underline{7.03} & \textbf{10.65} \\
\midrule
Ours         & --   & --   & \textbf{6.85} & \underline{11.22} \\
\bottomrule
\end{tabular*}
\caption{Comparisons with state-of-the-art methods on the SDD dataset. Text in bold numbers denotes the best result.}
\label{result2}
\end{table}

\subsection{Qualitative Evaluation}

To qualitatively evaluate the proposed model, several representative cases are selected and visualised. We compare our method with MID and Social-VAE across these scenarios.

\begin{figure}[t]
\centering
\includegraphics[scale= 0.077]{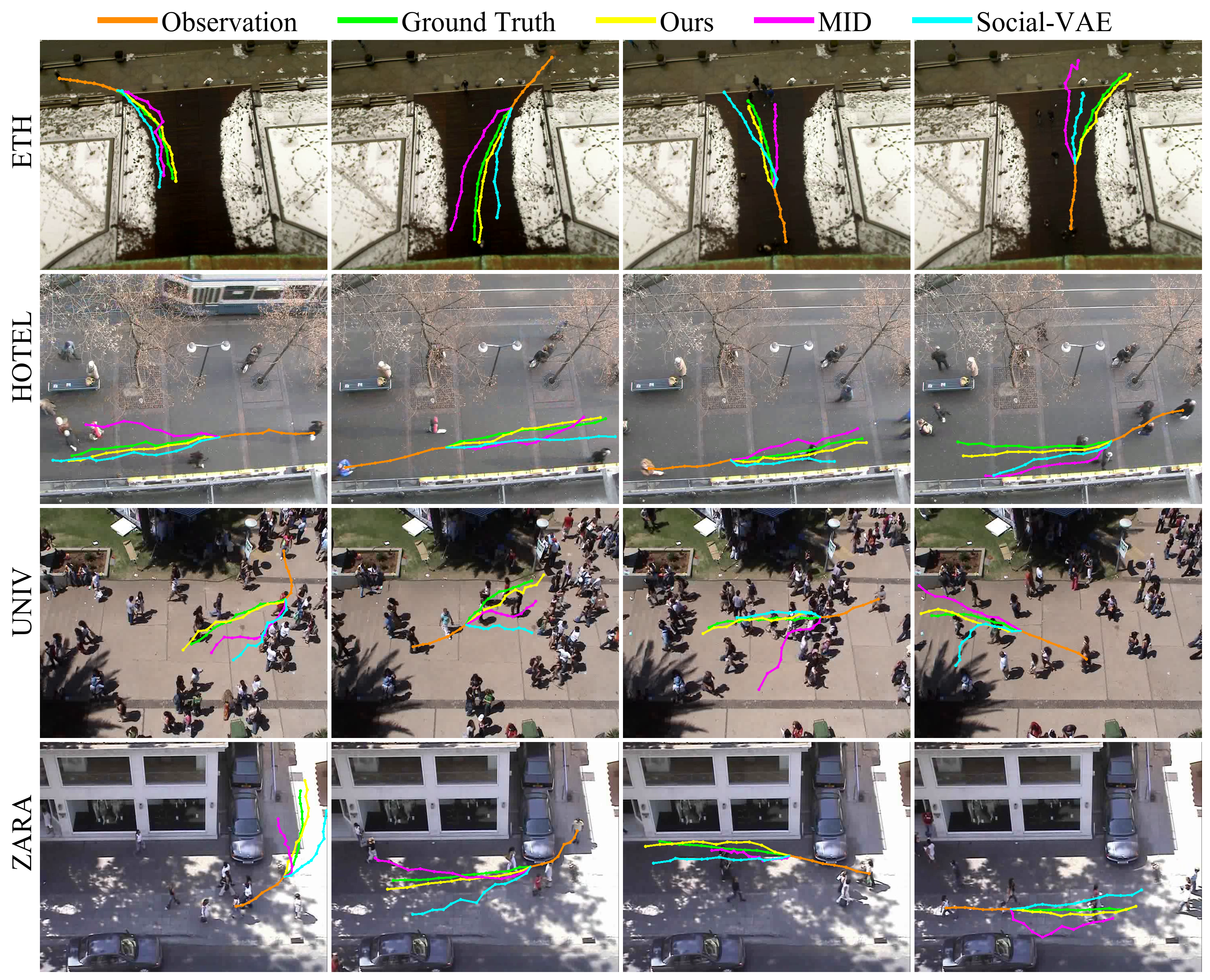}
\caption{Visualisation of prediction results on the ETH/UCY dataset. Our method (yellow) is compared with MID (purple) and Social-VAE (cyan) across four scenarios from ETH and UCY datasets.}
\label{vis1}
\end{figure}

\begin{figure}[t]
\centering
\includegraphics[scale= 0.08]{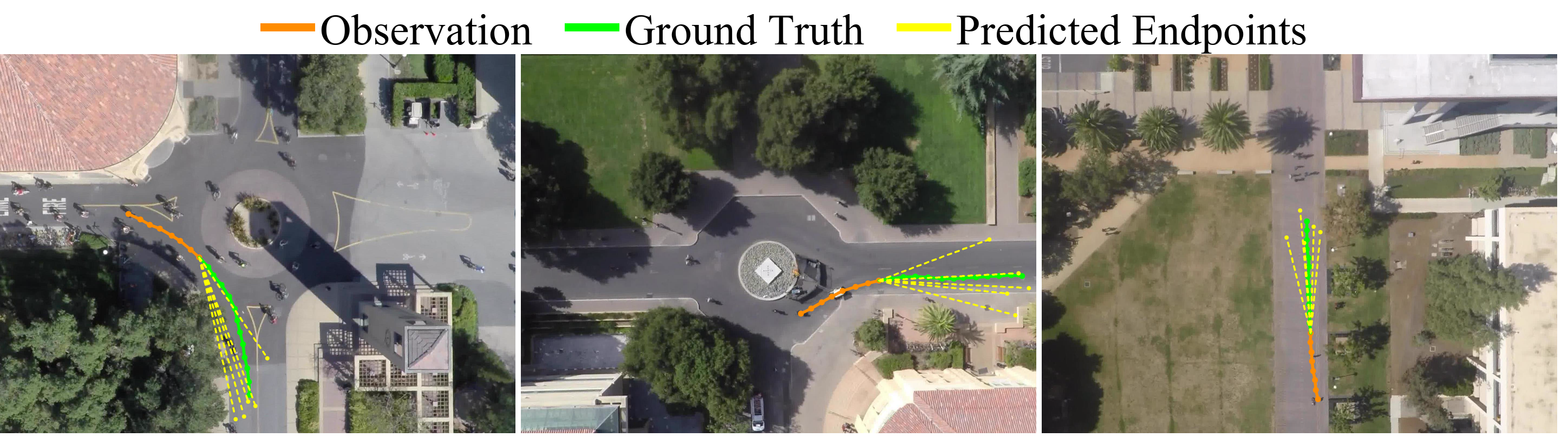}
\caption{Illustration of predicted endpoints on three scenarios from the SDD dataset.}
\label{vis3}
\end{figure}

As shown in Figure~\ref{vis1}, which visualizes four examples per dataset, our method consistently produces trajectories that align more closely with the ground truth than competing approaches. For instance, in the first two ETH cases involving subtle turns, all methods capture the motion trend, but ours more accurately follows the true path. In the first ZARA example, where a sharp turn occurs, only our method adapts well, producing a precise trajectory. Additionally, Figure~\ref{vis3} illustrates predicted endpoint candidates on the SDD dataset. While some predictions deviate, our model reliably includes candidates near the true goal, offering strong guidance for accurate future trajectory generation.


\subsection{Ablation Study}
To assess the impact of different model components, we conduct ablation studies on the following aspects.

Pedestrian motion intentions are modeled from both long-term and short-term perspectives, with a softmask mechanism and residual noise refinement jointly regulating their influence during the diffusion process. This ablation study evaluates the contribution of each component. As shown in Table~\ref{Ablation_int}, removing any of them degrades performance, highlighting their critical role in accurate trajectory prediction.

Similarly, an ablation study on the number of candidate endpoints $M$ was conducted, as shown in Table~\ref{Ablation_m}. Results indicate that prediction performance is sensitive to $M$, and the relationship is not linear. Overall performance across scenarios is maximized when $M$ is 5. A possible explanation is that too many candidates may dilute the model’s focus, making it harder to capture a coherent and interpretable intention distribution, while too few may limit its ability to represent the diversity of plausible intent hypotheses.

We further evaluate the impact of diffusion steps $K$ by varying it from 10 to 200, with detailed results shown in Table~\ref{Ablation_diffstep}. Overall prediction performance consistently peaks at $K=100$, where both ADE and FDE reach their lowest values. Too few steps may hinder the model’s ability to effectively capture fine-grained noise transitions, while too many make subtle differences between steps harder to distinguish, potentially exceeding the model’s representational capacity.

Lastly, we investigate refinement input and correction target. Using either noised state \(Y_k\) or predicted noise \(\epsilon_k\) as input, and refining either residual noise \(\Delta \epsilon_k\) or residual data \(\Delta Y_0\), we find correcting \(\Delta \epsilon_k\) yields better results (Table~\ref{Ablation_refine}). This suggests refining noise is more effective than adjusting data directly. Using noise state to correct predicted noise is also more accurate than using data state, likely due to noise space’s simpler, more continuous structure, while direct data adjustment is more complex and susceptible to errors.
\begin{table}[!h]
\centering
\setlength{\tabcolsep}{4.5pt} 
\fontsize{9}{10}\selectfont
\begin{tabular}{cccccc}
\toprule
 Long       & Short           & Softmask    & Refine          & ETH \& UCY  & SDD\\
\midrule
\checkmark  & \checkmark      & \checkmark  & \checkmark     & \textbf{0.19} / \textbf{0.31} & \textbf{6.85}/ \textbf{11.22}\\
\checkmark  & \checkmark      & \checkmark  &                & 0.23 / 0.35                   & 7.92/ 12.45\\
\checkmark  & \checkmark      &             & \checkmark     & 0.25 / 0.41                   & 9.25/ 13.82\\
\checkmark  &                 & \checkmark  & \checkmark     & 0.31 / 0.38                   & 8.96/ 13.12\\
            & \checkmark      & \checkmark  & \checkmark     & 0.29 / 0.36                   & 8.91/ 12.79\\
\bottomrule
\end{tabular}
\caption{Ablation study on model components on the ETH, UCY, and SDD. Bold is the best.}
\label{Ablation_int}
\end{table}
\begin{table}[!h]
\centering
\setlength{\tabcolsep}{3pt}  
\fontsize{9}{10}\selectfont
\begin{tabular}{l|cccccc}
\toprule
 M & $1$ & $3$ & $5$ & $10$ & $20$\\
\midrule
ETH     & 0.39/0.61 & 0.36/0.56 & 0.34/\textbf{0.52} & \textbf{0.33}/\textbf{0.52}  & 0.40/0.59 \\
HOTEL   & 0.18/0.31 & \textbf{0.15}/0.28 & \textbf{0.15}/\textbf{0.24} & 0.18/0.25  & 0.21/0.26 \\
UNIV    & 0.25/0.39 & 0.21/\textbf{0.35} & \textbf{0.20}/0.36 & \textbf{0.20}/0.41  & 0.23/0.42 \\
ZARA1   & 0.20/0.29 & 0.18/0.27 & \textbf{0.15}/\textbf{0.24} & 0.16/0.30  & 0.24/0.32 \\
ZARA2   & 0.16/0.29 & \textbf{0.11}/0.24 & \textbf{0.11}/\textbf{0.20} & 0.17/\textbf{0.20}  & 0.29/0.32 \\
\midrule
AVG     & 0.24/0.38 & 0.20/0.34 & \textbf{0.19}/\textbf{0.31} & 0.21/0.34  & 0.27/0.38\\
\bottomrule
\end{tabular}
\caption{Ablation study on the number of endpoint candidates $M$ on the ETH and UCY datasets. Bold is the best.}
\label{Ablation_m}
\end{table}
\begin{table}[!h]
\centering
\setlength{\tabcolsep}{3pt}  
\fontsize{9}{10}\selectfont
\begin{tabular}{l|cccccc}
\toprule
 K & $10$ & $50$ & $100$ & $150$ & $200$\\
\midrule
ETH     & 0.37/0.58 & \textbf{0.34}/0.53 & \textbf{0.34}/\textbf{0.52} & 0.36/0.60  & 0.39/0.65 \\
HOTEL   & 0.16/0.29 & 0.16/\textbf{0.23} & \textbf{0.15}/0.24 & 0.18/0.25  & 0.21/0.27 \\
UNIV    & 0.23/0.38 & \textbf{0.20}/0.38 & \textbf{0.20}/\textbf{0.36} & 0.23/\textbf{0.36}  & 0.23/0.42 \\
ZARA1   & 0.19/0.29 & 0.17/0.26 & 0.15/\textbf{0.24} & \textbf{0.14}/0.25  & 0.16/0.25 \\
ZARA2   & 0.13/0.22 & \textbf{0.11}/0.22 & \textbf{0.11}/\textbf{0.20} & 0.12/0.23  & 0.14/0.23 \\
\hline
AVG     & 0.22/0.35 & 0.20/0.32 & \textbf{0.19}/\textbf{0.31} & 0.21/0.34  & 0.23/0.36\\
\bottomrule
\end{tabular}
\caption{Ablation study on diffusion step ${K}$ on the ETH and UCY datasets. Bold is the best.}
\label{Ablation_diffstep}
\end{table}
\begin{table}[!h]
\centering
\setlength{\tabcolsep}{8pt} 
\fontsize{10}{10}\selectfont
\begin{tabular}{clcc}
\toprule
 In                 & Out                       & ETH \& UCY                    & SDD\\
\midrule
${\epsilon}_k$      & $\Delta {\epsilon_k}$     & \textbf{0.19} / \textbf{0.31} & \textbf{6.85}/ \textbf{11.22}\\
${\epsilon}_k$      & $\Delta {Y_0}$            & 0.39 / 0.48                   & 9.97/ 15.15\\
$Y_k$               & $\Delta {\epsilon_k}$     & 0.24 / 0.39                   & 7.65/ 12.82\\
$Y_k$               & $\Delta {Y_0}$            & 0.41 / 0.50                   & 9.86/ 14.22\\
\bottomrule
\end{tabular}
\caption{Ablation study of the refinement module on the ETH, UCY, and SDD datasets.}
\label{Ablation_refine}
\end{table}

\section{Conclusion} In this work, we propose a diffusion-based trajectory prediction framework enhanced by long- and short-term intentions. Short-term motion is represented in polar coordinates for precise local modeling, while long-term intention is captured via predicted endpoints providing global cues. Softmask-based classifier-free guidance and residual noise estimation improve generation quality by enhancing error alignment during denoising. Experiments demonstrate competitive performance against state-of-the-art methods.

\bibliography{aaai2026}
\end{document}